\documentclass[10pt,twocolumn,letterpaper]{article}

\usepackage[pagenumbers]{wacv} 

\newcommand{\red}[1]{{\color{red}#1}}

\usepackage{graphicx}
\graphicspath{{./images/}}
\usepackage{multirow}
\usepackage{amsmath}
\usepackage{tabularx, booktabs}

\newcolumntype{Y}{>{\centering\arraybackslash}X}

\newcommand\clearrow{\global\let\rowmac\relax}
\clearrow

\definecolor{wacvblue}{rgb}{0.21,0.49,0.74}
\usepackage[pagebackref,breaklinks,colorlinks,allcolors=wacvblue]{hyperref}

\def\wacvPaperID{1142}
\def\confName{WACV}
\def\confYear{2026}

\usepackage{fancyhdr}
\title{Flood-LDM: Generalizable Latent Diffusion Models for rapid and accurate zero-shot High-Resolution Flood Mapping}

\author{Sun Han Neo\textsuperscript{1}, Sachith Seneviratne\textsuperscript{2}, Herath Mudiyanselage Viraj Vidura Herath\textsuperscript{3},\\
Abhishek Saha\textsuperscript{4}, Sanka Rasnayaka\textsuperscript{1}, Lucy Amanda Marshall\textsuperscript{3}\smallskip\\
\textsuperscript{1}Department of Computer Science, School of Computing, National University of Singapore\\
\textsuperscript{2}Transport, Health and Urban Systems Research Lab, Melbourne School of Design, University of Melbourne\\
\textsuperscript{3}School of Civil Engineering, Faculty of Engineering, University of Sydney\\
\textsuperscript{4}Delft Institute of Applied Mathematics, Delft University of Technology\\
{\tt\small \shortstack{neosunhan@u.nus.edu, sachith.seneviratne@unimelb.edu.au, viraj.herath@sydney.edu.au,\\abhishek@h2i.sg, sanka@nus.edu.sg, lucy.marshall@sydney.edu.au}}
}

\begin{document}
\maketitle
\thispagestyle{fancy}

\begin{abstract}
Flood prediction is critical for emergency planning and response to mitigate human and economic losses. Traditional physics-based hydrodynamic models generate high-resolution flood maps using numerical methods requiring fine-grid discretization; which are computationally intensive and impractical for real-time large-scale applications. While recent studies have applied convolutional neural networks for flood map super-resolution with good accuracy and speed, they suffer from limited generalizability to unseen areas. In this paper, we propose a novel approach that leverages latent diffusion models to perform super-resolution on coarse-grid flood maps, with the objective of achieving the accuracy of fine-grid flood maps while significantly reducing inference time. Experimental results demonstrate that latent diffusion models substantially decrease the computational time required to produce high-fidelity flood maps without compromising on accuracy, enabling their use in real-time flood risk management. Moreover, diffusion models exhibit superior generalizability across different physical locations, with transfer learning further accelerating adaptation to new geographic regions. Our approach also incorporates physics-informed inputs, addressing the common limitation of black-box behavior in machine learning, thereby enhancing interpretability. Code is available at \hyperlink{https://github.com/neosunhan/flood-diff}{https://github.com/neosunhan/flood-diff}.
\end{abstract}

\section{Introduction}
\label{sec:introduction}
Floods represent one of the most frequent and destructive natural disasters worldwide, causing widespread loss of life and property \cite{gar2022}. Accurate and timely flood prediction is critical for emergency planning and response, enabling authorities to issue warnings, allocate resources, and execute evacuation plans to mitigate human and economic losses.


Central to flood prediction efforts are flood maps, which visualize the spatial distribution of water depth across terrain. They are used to identify flood-prone regions and predict the extent and depth of water inundation. Flood maps also provide crucial information for designing flood defences and evacuation routes. Effective flood mapping requires a balance between prediction accuracy, computational speed, generalizability across diverse regions, and interpretability of the underlying physical dynamics.

Traditionally, flood maps are produced using physics-based hydrodynamic models. These models numerically solve the governing physical equations on discretized terrain grids, providing accurate and interpretable results. Finer grids yield more detailed forecasts \cite{jodhani_23}, but at the cost of significantly higher computational demands \cite{jafarzadegan_23}, which makes them impractical for real-time applications. To overcome these computational constraints, data-driven super-resolution methods, primarily based on Convolutional Neural Networks (CNNs), have been developed to upsample coarse hydrodynamic outputs \cite{he_23,choi_25}. These CNN-based approaches deliver rapid, high-fidelity predictions \cite{yin_24}, but frequently overfit to specific catchments, or physical locations, and struggle to generalize to new regions \cite{song_25,sgunet}.

\begin{figure*}[t]
    \centering
    \includegraphics[width=0.9\linewidth]{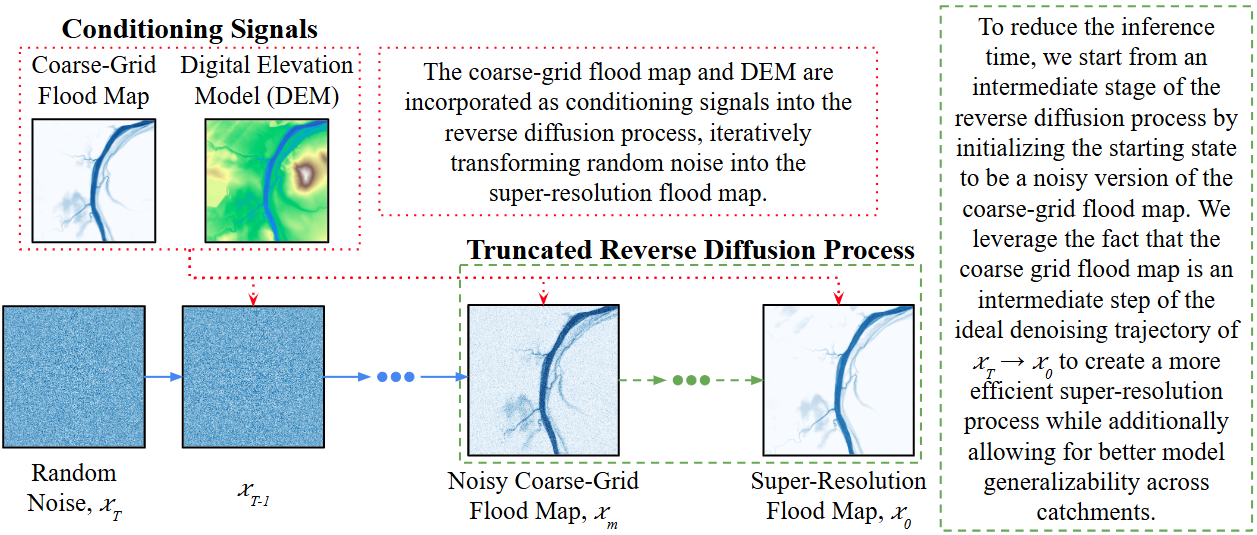}
    \caption{Overview of our proposed approach.}
    \label{fig:proposed_approach}
\end{figure*}

To address these limitations, we propose the first diffusion-based framework for flood-map super-resolution. As illustrated in Figure \ref{fig:proposed_approach}, our approach uses a conditional diffusion model (DM) to iteratively refine coarse-grid simulations into high-fidelity flood maps through a truncated denoising process. The DM incorporates the coarse-grid flood map and Digital Elevation Model (DEM) as physics-informed conditioning signals, functioning as a surrogate model that achieves accuracy comparable to fine-grid hydrodynamic models without the associated long computational time. This hybrid approach merges the reliability and interpretability of physics-based simulations with the generalization strengths of probabilistic generative modeling.

Our contributions are summarized as follows:
\begin{itemize}
    \item We introduce the first diffusion-based framework for flood-map super-resolution, motivated by DMs’ enhanced generalizability over CNN-based methods that often overfit to training catchments.
    \item We ensure the real-world applicability of our approach by supporting real-time flood forecasting via efficient sampling strategies for rapid inference. In addition, the use of physics-informed coarse-grid inputs serves to preserve physical interpretability, ensuring the model remains reliable for trusted use in critical applications such as disaster response and flood risk management.
    \item We demonstrate the model’s ability to generalize to unseen catchments without compromising high-resolution accuracy. This is crucial in situations where time or data availability constraints prevent extensive retraining for a newly observed area.
\end{itemize}

\section{Related Work}
\label{sec:related}
\subsection{Flood Mapping Techniques}

\begin{figure}[t]
    \centering
    \includegraphics[width=\linewidth]{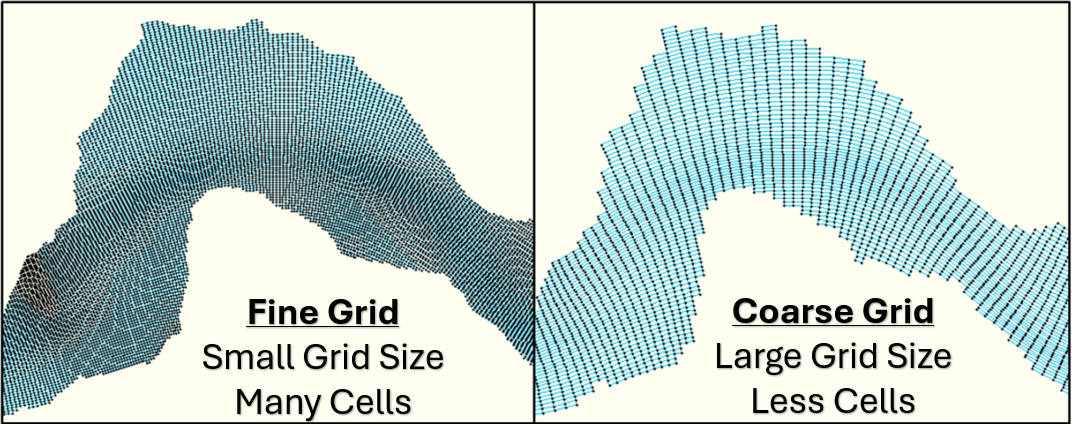}
    \caption{Comparison of fine- vs coarse-grid flood maps \protect\cite{bomers_19}. Coarse grids use fewer, larger cells, enabling faster computation but reducing spatial detail, whereas fine grids use more, smaller cells, providing greater accuracy at higher computational cost.}
\label{fig:grid_size_comparison}
\end{figure}

Traditionally, flood maps are produced by physics-based hydrodynamic models, which rely on the principles of fluid dynamics to provide a physically interpretable and accurate representation of flood behaviour \cite{jodhani_23} with the downside of being computationally intensive \cite{jafarzadegan_23}. Hydrodynamic models typically discretize the target area into a structured or unstructured mesh \cite{mudashiru_21}, before numerically solving the two-dimensional Saint-Venant equations (also known as the shallow water equations) to compute water depth within each grid cell and generate the flood map.  \Cref{fig:grid_size_comparison} shows the effect of the grid size, a key hyperparameter in this modeling process that determines the tradeoff between computational efficiency and accuracy. Broadly speaking, fine grids increase spatial detail but sacrifice computational speed. 

The large computational time of traditional hydrodynamic models led to the development of surrogate models, which try to achieve comparable levels of accuracy compared to hydrodynamic models but in a more computationally efficient way \cite{zahura_22}. These models involve data-driven approaches which forego solving hydrodynamic equations in favour of using machine learning algorithms to predict flood extents \cite{kumar_23}. While their performance is upper-bounded by the hydrodynamic models used to create the training data, their speed makes them a valuable tool for real-time flood forecasting \cite{karim_23}. In recent years, deep learning has gained popularity in flood mapping as a faster and more flexible alternative to traditional models \cite{bentivoglio_22}. CNNs are particularly effective for processing spatial data, such as satellite images or DEMs \cite{tavus_22}, while Recurrent Neural Networks (RNNs) are suitable for handling temporal data like rainfall sequences \cite{xiang_20}. These models are typically faster than physics-based models but often struggle with generalizability when applied to new geographic areas, as they may overfit to the specific features of the training dataset \cite{mosavi_18}.

Another challenge with data-driven models, including deep learning approaches, is their low interpretability \cite{murdoch_19}. Despite their speed, these models are often seen as ``black-box'' solutions, which can hinder their acceptance in critical decision-making contexts. To address this, researchers are exploring hybrid models, also known as physics-guided models, that integrate the strengths of both hydrodynamic and data-driven approaches \cite{bhattarai_24}. Physics-guided models offer the interpretability and reliability of physics-based models \cite{fraehr_24} alongside the speed and efficiency of data-driven approaches \cite{yang_24}. In recent years, super-resolution of flood maps has emerged as a practical solution for balancing the trade-off between computational efficiency and model interpretability in flood forecasting applications \cite{sgunet}. This approach involves generating a coarse-grid flood map using a physics-based hydrodynamic model and subsequently enhancing its accuracy through a learned super-resolution model \cite{yin_24}. Using the coarse-grid hydrodynamic simulation as the core input, these methods preserve the physics-guided nature of the flood prediction process, ensuring that the resulting high-resolution flood maps remain grounded in established principles of fluid dynamics \cite{sgunet}. This is crucial in operational flood risk management, where interpretability and alignment with physical laws are critical to building trust in automated forecasting systems \cite{ding_20}.

To date, most research on flood map super-resolution has focused on CNNs \cite{sgunet,he_23,yin_24,song_25,choi_25}. These models have demonstrated strong performance in accurately predicting high-resolution flood maps from low-resolution inputs, while achieving rapid inference speeds by producing results in a single forward pass \cite{yin_24}. The U-Net architecture is widely used within CNN-based flood map super-resolution models \cite{sgunet,he_23,yin_24,song_25}. Originally developed for biomedical image segmentation tasks \cite{ronneberger_15}, it has demonstrated considerable success when adapted for image super-resolution applications due to its encoder-decoder structure and skip connections, which enable the preservation of spatial information across multiple resolutions \cite{hu_19}. The U-Net's ability to efficiently capture local and regional flood patterns has led to strong performance when applied to catchments represented within the training data \cite{sgunet}. 

However, a notable limitation of the U-Net, and CNN-based models in general, is their lack of generalizability \cite{song_25,yin_24}. These models often struggle when applied to catchments or flood events outside their training distribution, particularly in regions with differing hydrological, topographical, or climatic conditions \cite{he_23,choi_25}. Consequently, we turn to exploring alternative architectures, such as DMs, to overcome these generalization challenges while maintaining the computational efficiency and accuracy necessary for large-scale, real-time flood forecasting.

\subsection{Diffusion Models}
In recent years, DM has emerged as one of the most promising approaches in the field of natural image super-resolution \cite{moser_24}. These models consist of two components: the forward diffusion process $q$ (\cref{eqn:forward_diffusion}) that iteratively adds noise to the original image $x_0$ over a series of $T$ timesteps (\( t \in \{1, 2, \dots, T\} \))  and the reverse diffusion process $p_\theta$ (\cref{eqn:reverse_diffusion}) that iteratively removes noise, starting from random noise $x_T$ and moving back to an estimate of $x_0$. In \cref{eqn:forward_diffusion}, $\alpha_t$ controls the variance of the Gaussian noise added at each timestep $t$. In \cref{eqn:reverse_diffusion}, $\mu_\theta(x_t, t)$ is the learned mean and $\Sigma_\theta(x_t, t)$ is the learned covariance matrix of the reverse process. \cref{eqn:forward_closed_form} provides a closed-form expression for the marginal distribution $q(x_t \mid x_0)$, where $\gamma_t = \prod_{s=1}^{t} (1 - \alpha_s)$ represents the cumulative noise schedule. The number of timesteps $T$ is a critical hyperparameter for both processes, as larger values can yield more accurate reconstructions but also increase computation time.

\begin{equation}
    \label{eqn:forward_diffusion}
    q(x_t | x_{t-1}) = \mathcal{N}(x_t; \sqrt{1 - \alpha_t} \, x_{t-1}, \alpha_t \mathbf{I})
\end{equation}
\begin{equation}
    \label{eqn:reverse_diffusion}
    p_\theta(x_{t-1} | x_t) = \mathcal{N}(x_{t-1}; \mu_\theta(x_t, t), \Sigma_\theta(x_t, t))
\end{equation}
\begin{equation}
    \label{eqn:forward_closed_form}
    q(x_t | x_0) = \mathcal{N}(x_t; \sqrt{\gamma_t} \, x_0, (1 - \gamma_t) \mathbf{I})
\end{equation}

DMs have demonstrated state-of-the-art performance in generating high-quality, diverse image samples, surpassing traditional methods in many benchmark image restoration tasks \cite{palette,sr3,improved_ddpm}. However, despite their success in the natural image domain, there has been little to no research exploring the use of DMs for super-resolution of coarse-grid flood maps. This represents a significant gap in the current literature, as flood mapping applications could greatly benefit from the generalization capabilities and high-fidelity outputs that DMs offer \cite{cache_24}. A key advantage of DMs lies in their ability to generate diverse and realistic outputs due to their probabilistic formulation \cite{kazerouni_23}. This makes them well-suited for applications such as flood mapping, where models must often generalize to new catchment areas or unfamiliar flood scenarios without extensive retraining. 

However, one of the main limitations of DMs is their prolonged inference time compared to CNNs. While the forward diffusion process benefits from a closed-form solution (\cref{eqn:forward_closed_form}) that allows for direct single-step sampling of $x_t$, no such equivalent exists for the reverse diffusion process, necessitating multiple sequential passes through the model \cite{croitoru_23}. To address this issue, one notable advancement is the latent diffusion model (LDM) \cite{rombach_22}, which performs the diffusion process in a lower-dimensional latent space and thus significantly decreases model complexity, leading to faster training and inference. Typically, a variational autoencoder is used to encode and decode the input image at the start and end of the forward and reverse diffusion processes.

\begin{table*}[t]
    \centering
    \begin{tabular}{ccccccccccc}
        \toprule
        \multirow{2}{*}{Catchment} & \multirow{2}{*}{\shortstack{No. of\\Patches}} & \multicolumn{2}{c}{No. of Images} & \multirow{2}{*}{\shortstack{Max Depth\\(cm)}} & \multicolumn{2}{c}{No. of Cells} & \multirow{2}{*}{\shortstack{Upscale\\Factor}} & \multirow{2}{*}{\shortstack{Area\\(km$^2$)}} & \multirow{2}{*}{\shortstack{Mapping\\Interval}} & \multirow{2}{*}{\shortstack{Resolution}}\\
        \cmidrule(lr){3-4} \cmidrule(lr){6-7}
        & & Train & Test & & CG & FG & & & \\
        \midrule
        1 & 21 & 35280 & 6048 & 814 & 2612 & 71487 & 27.4 & 1119 & 30 min & 5m $\times$ 5m \\
        2 & 40 & 90240 & 17280 & 760 & 3421 & 94780 & 27.7 & 3059 & 30 min & 5m $\times$ 5m \\
        3 & 100 & 160200 & 49400 & 1197 & 4601 & 395792 & 86 & 617 & 6 hours & 10m $\times$ 10m \\
        \bottomrule
    \end{tabular}
    \caption{Catchment statistics. The mapping interval denotes the time step between successive flood maps. CG and FG columns show the number of coarse- and fine-grid cells, with their ratio as the upscale factor. Each catchment was divided into overlapping 512 × 512 patches. Three to five rainfall events were used for training and one for testing. Although CG and FG maps share the DEM resolution, CG simulations have fewer computational cells.}
    \label{tab:data_statistics}
\end{table*}

\begin{figure*}[t]
    \centering
    \includegraphics[width=\linewidth]{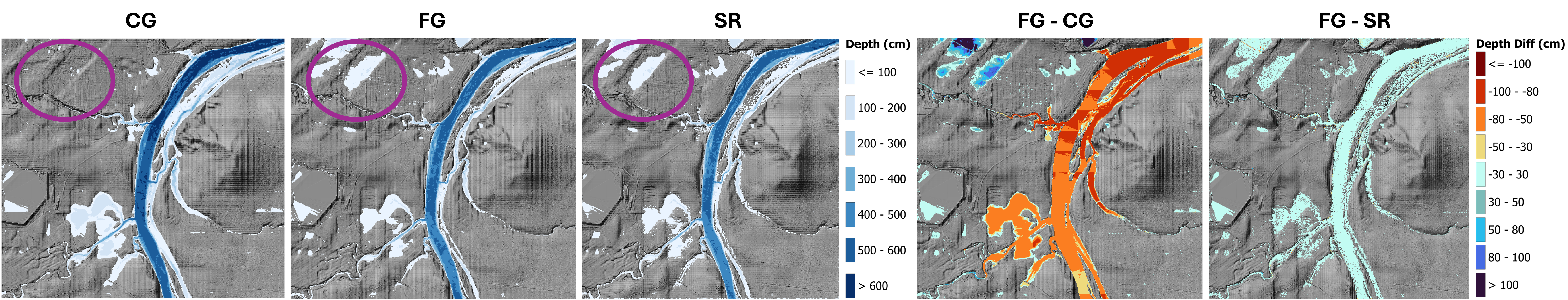}
    \caption{Comparison of coarse-grid (CG), fine-grid (FG), and super-resolution (SR) flood maps for a segment of Catchment 1. The purple circle marks a flooded area missed by the CG but successfully recovered by the SR model. The right panels show depth differences (FG – CG and FG – SR), where negative values in FG – CG indicate CG overestimation. The SR map reduces these errors, closely matching the FG at lower cost, while preserving flood contours and providing sharp predictions without distorting inundation boundaries.}
    \label{fig:flood_map_comparison}
\end{figure*}

\section{Methodology}
\label{sec:methodology}
\subsection{Data}
\label{sec:data}
Data from three Australian watersheds, Wollombi, Chowilla, and Burnett River, was used to evaluate the performance of the proposed approach. These regions are referred to as Catchment 1, 2 and 3 respectively in the rest of the paper. These catchments were selected due to their diverse hydrodynamic characteristics: Catchment 1 is steep with short rainfall- and inflow-driven floods, Catchment 2 is flat with prolonged inflow-driven floods, and Catchment 3 is steep with compounding inland and coastal influences, making it the most complex. For each of the three catchments, the HEC-RAS 2D hydrodynamic model \cite{hecras} was used to generate coarse-grid and fine-grid flood maps at regular time intervals during rain events. The flood maps were subsequently divided into overlapping images. The final number of images in the train and test sets of each catchment can be viewed in \cref{tab:data_statistics} along with other details. Additional statistics for each catchment are available in the supplementary material. For further details on the training data and catchment flood dynamics, please refer to \cite{sgunet}.

As seen in \cref{fig:proposed_approach}, the coarse-grid flood maps and their corresponding DEMs were provided as conditioning signals to the DM. The DEM is a representation of the bare-earth terrain surface, excluding vegetation, buildings, and other surface features, and provides crucial topographic information to the model. The fine-grid flood maps, generated by the hydrodynamic model, served as ground-truth references for evaluating the DM's output. \Cref{fig:flood_map_comparison} shows an example of the coarse-grid, fine-grid and super-resolution flood maps for the same area. In each catchment, the DEM was cropped to match the spatial extent of the flood map images.


\begin{table*}[t]
    \centering
    \begin{tabular}{cccccccccccc}
        \toprule
        & \multicolumn{3}{c}{Catchment 1} & \multicolumn{3}{c}{Catchment 2} & \multicolumn{3}{c}{Catchment 3} & \\
        \cmidrule(lr){2-4} \cmidrule(lr){5-7} \cmidrule(lr){8-10} 
        Model & \shortstack{CG-FG\\MSE} & \shortstack{SR-FG\\MSE} & \shortstack{\%\\change} & \shortstack{CG-FG\\MSE} & \shortstack{SR-FG\\MSE} & \shortstack{\%\\change} & \shortstack{CG-FG\\MSE} & \shortstack{SR-FG\\MSE} & \shortstack{\%\\change} & \shortstack{Variance in\\\% change $\downarrow$} \\
        \midrule
        SGUnet \cite{sgunet} & 344.2 & 28.1 & -91.84 & 5957.4 & 725.8 & -87.82 & 158.7 & 55.8 & -64.82 & 41.69 \\
        DM (ours) & \textbf{344.2} & \textbf{21.6} & \textbf{-93.71} & 5957.4 & 2022.0 & -66.06 & \textbf{158.7} & \textbf{14.8} & \textbf{-90.70} & 153.41 \\
        LDM (ours) & 344.2 & 33.7 & -90.20 & \textbf{5957.4} & \textbf{723.0} & \textbf{-87.86} & 158.7 & 17.4 & -89.07 & \textbf{0.91} \\
        \bottomrule
    \end{tabular}
    \caption{Decrease in MSE for different models in each catchment. All raw MSE values are in cm$^2$. The LDM exhibits the lowest variance in performance across catchments and displays the best overall performance compared to the other models due to its consistency.}
    \label{tab:mse}
\end{table*}

\subsection{Model Architecture}
A DM architecture was developed based on the popular SR3 architecture \cite{sr3}. While preserving the core U-Net architecture of the SR3 framework, modifications were made to accommodate the 512 $\times$ 512 input dimensions used in this study. Hyperparameters, such as the denoising schedule and attention layers, were retained at their default settings as outlined in the SR3 paper. Additionally, the DEM was incorporated as a conditioning signal via channel concatenation to enhance model performance. The DM used a 5-layer U-Net architecture with approximately 317 million parameters, and was trained for 400,000 steps on each of the three catchment datasets. A linear schedule with 1000 timesteps was used in the training phase of the DM. Pixel values were normalized to $[0, 1]$ and L2 loss was used to train the model.

To provide a baseline comparison, a fully convolutional model following the current state-of-the-art SGUnet architecture \cite{sgunet} was used. This model was trained on Catchments 1, 2, and 3 for 75, 100 and 15 epochs respectively.

Finally, a LDM architecture was developed based on the DDPM framework \cite{ddpm}. A pretrained variational autoencoder was used to transform the original 512 $\times$ 512 single-channel image into a 64 $\times$ 64 image with 4 channels. The DEM also experienced a transformation with the same dimensions before being incorporated into the LDM as an additional conditioning signal via channel concatenation. The LDM used a 5-layer U-Net architecture with approximately 156 million parameters, and was trained for 300,000 - 400,000 steps on each of the three catchment datasets. A linear schedule with 1000 timesteps was used in the training phase of the LDM. Pixel values were normalized to $[-1, 1]$ and L2 loss was used to train the model.

Model evaluation was performed on a randomly sampled subset of 1,000 images from each catchment’s test dataset, using the same subset across all models for fair comparison.

\section{Experimental Results}
\label{sec:results}
To evaluate the model performance, the mean squared error (MSE) metric is used with the following equation:
\begin{equation}
    \text{MSE} = \frac{1}{n} \sum_{i=1}^n (y_i - \hat{y}_i)^2
\end{equation}
where $n$ is the total number of pixels in the flood maps, $y_i$ is the water depth of the $i$-th pixel in the ground truth fine-grid flood maps, and $\hat{y}_i$ is the predicted water depth of the $i$-th pixel in the output flood maps of the model.

As previously outlined, the hydrodynamic model will generate both the coarse-grid and fine-grid flood maps, and the proposed approach converts the coarse-grid flood map to a super-resolution flood map. To evaluate the model’s effectiveness, we compare two values: the initial MSE between the coarse-grid and fine-grid flood maps, and the final MSE between the super-resolution model output and the fine-grid flood map. These are referred to as the CG-FG MSE and SR-FG MSE, respectively, throughout this paper.

Model performance is considered satisfactory when there is a significant reduction from the CG-FG MSE to the SR-FG MSE, indicating that the super-resolution model has effectively corrected the inaccuracies present in the coarse-grid flood map. Given the variability in absolute MSE values across all catchments, the evaluation focuses primarily on the percentage change in MSE to enable a consistent and meaningful comparison across different geographic areas.

\subsection{Model Comparison}
For each of the three catchments, all three model architectures were trained and evaluated on their respective datasets, with results presented in \cref{tab:mse}. While all models achieved substantial reductions in MSE, the LDM demonstrated the smallest variance in percentage change across catchments, indicating greater stability and more consistent performance across diverse scenarios. It is important to note that these results were obtained under an idealised condition where models were trained on a large and diverse dataset from the target catchment, a scenario that does not reflect operational realities. In practical applications, it is often infeasible to assemble substantial training datasets for new catchments within a limited timeframe, largely due to the computational expense associated with generating fine-grid ground-truth flood maps. As such, it is essential for models to learn a feature space that is generalizable and capable of delivering reliable performance in zero-shot scenarios, where no retraining is performed on data from the new location.

\begin{table*}[t]
    \centering
    \begin{tabular}{cccccccc}
        \toprule
        & & \multicolumn{3}{c}{Test Catchment 1} & \multicolumn{3}{c}{Test Catchment 2} \\
        \cmidrule(lr){3-5} \cmidrule(lr){6-8} 
        \shortstack{Training\\ Catchment} & Model & \shortstack{CG-FG MSE\\ (cm$^2$)} & \shortstack{SR-FG MSE\\ (cm$^2$)} & \% change $\downarrow$ & \shortstack{CG-FG MSE\\ (cm$^2$)} & \shortstack{SR-FG MSE\\ (cm$^2$)} & \% change $\downarrow$ \\
        \midrule
        \multirow{3}{*}{3} & SGUnet \cite{sgunet} & 344.2 & 2642.6 & +667.84 & 5957.4 & 7829.9 & +31.43 \\
        & DM (ours) & 344.2 & 842.0 & +144.67 & \textbf{5957.4} & \textbf{3329.7} & \textbf{-44.11} \\
        & LDM (ours) & \textbf{344.2} & \textbf{345.0} & \textbf{+0.26} & 5957.4 & 4602.4 & -22.75 \\
        \bottomrule
    \end{tabular}
    \caption{All models were trained on Catchment 3 and subsequently evaluated on the unseen Catchments 1 and 2. The diffusion-based architectures significantly outperform the SGUnet, showcasing their increased generalizability in zero-shot settings over CNN-based models.}
    \label{tab:chowilla_generalizability}
\end{table*}

The generalizability of the models is evaluated in \cref{tab:chowilla_generalizability}, which reports performance on Catchments 1 and 2 for models trained exclusively on the Catchment 3 dataset. Consistent trends were observed for models trained on the other two catchments and recorded in the Supplementary Material. The results indicate that both the standard DM and the LDM exhibit markedly better generalization capabilities than the fully convolutional SGUnet baseline. Notably, the DMs trained on Catchment 3 achieved a significant reduction in MSE when applied to Catchment 2, despite the absence of any data from Catchment 2 in their training. In contrast, the SGUnet model exhibited increased MSE under these conditions, underscoring its limited capacity to generalize effectively to unseen geographical regions.

\begin{table}[t]
    \centering
    \begin{tabular}{cccc}
        \toprule
        Model & \shortstack{\% change\\in MSE\\on seen\\catchments $\downarrow$} & \shortstack{\% change\\ in MSE\\on unseen\\catchment $\downarrow$} & \shortstack{Inference\\Time $\downarrow$} \\
        \midrule
        SGUnet \cite{sgunet} & -81.49 & +990.80 & \textbf{0:00:27} \\
        DM (ours) & -83.49 & \textbf{+143.29} & 11:49:24 \\
        LDM (ours) & \textbf{-89.04} & +362.55 & 0:03:04 \\
        \bottomrule
    \end{tabular}
    \caption{Summary of model performance. The second column reports averages when training and testing on the same catchment; the third column shows averages when training on one catchment and testing on the other two (generalizability). The last column gives average inference time on 1000 images. All models perform comparably on same-catchment data, with LDM most consistent across catchments. While the standard DM generalizes best, its inference time is impractically high. Overall, LDM delivers the highest accuracy, strong generalizability, and acceptable runtime, making it the best-performing model.}
    \label{tab:results_summary}
\end{table}

Inference time comparisons between the three models revealed considerable disparities. \Cref{tab:results_summary} presents the inference durations for each model when applied to a test set of 1000 images, alongside a consolidated summary of their performance on both seen and unseen catchments. SGUnet achieved the fastest inference time, while the LDM and standard DM were approximately 10 times and 1000 times slower respectively. Nevertheless, the LDM consistently achieved the best overall performance across all evaluation metrics, being significantly more generalizable than the SGUnet baseline while still maintaining comparable levels of computational efficiency. Its inference speed was improved through a series of optimizations incorporated into the model pipeline, which are discussed in \cref{sec:inference_time_reduction}.

\subsection{Inference Time Reduction}
\label{sec:inference_time_reduction}
\begin{table*}
    \centering
    \begin{tabular}{c|c|c||ccc|c}
        \toprule
        Catchment & \shortstack{Inference Startpoint} & \shortstack{Inference Timesteps} & \shortstack{CG-FG MSE} & \shortstack{SR-FG MSE} & \% change $\downarrow$ & \shortstack{Time Taken $\downarrow$} \\
        \midrule
        \multirow{2}{*}{1} & \shortstack{Random Noise} & 1000 & 344.2 & 33.7 & -90.20 & 8:46 \\
        & \shortstack{Noisy CG Flood Map} & 50 & 344.2 & 33.8 & -90.17 & 1:42 \\
        \hline
        \multirow{2}{*}{2} & \shortstack{Random Noise} & 1000 & 5957.4 & 723.0 & -87.86 & 8:46 \\
        & \shortstack{Noisy CG Flood Map} & 500 & 5957.4 & 669.0 & -88.77 & 5:04 \\
        \hline
        \multirow{2}{*}{3} & \shortstack{Random Noise} & 1000 & 158.7 & 17.4 & -89.07 & 8:46 \\
        & \shortstack{Noisy CG Flood Map} & 150 & 158.7 & 17.7 & -88.83 & 2:27 \\
        \bottomrule
    \end{tabular}
    \caption{Comparison of LDM performance and inference time starting from random noise versus the noisy coarse-grid flood map. Starting from the noisy coarse-grid map (truncated reverse diffusion process in \cref{fig:proposed_approach}) greatly reduces inference timesteps, significantly speeding up computation while keeping \%MSE change below 1\% in all catchments.}
    \label{tab:inference_reduction}
\end{table*}

\Cref{tab:inference_reduction} summarizes the inference speed‑up achieved by the LDM in each catchment. Initially, inference on a test set of 1000 images required 8 minutes and 46 seconds. Two optimisations were employed to reduce this time:
\begin{enumerate}
    \item \textbf{Reduced inference timesteps:} By decoupling the noise schedules for training and inference, the model can be trained with a full set of timesteps but evaluated with fewer timesteps during inference. Although this typically introduces a minor increase in SR‑FG MSE, the reduction in timesteps yields significant speed gains.
    \item \textbf{Alternative initialization via noisy coarse-grid flood map:} As seen in \cref{fig:proposed_approach}, the reverse diffusion process typically begins from random noise $x_T$ and iteratively denoises the image to produce the final output $x_0$. Since the coarse-grid flood map can be viewed as a less-accurate version of the super-resolution flood map, we hypothesize that there is an intermediate output $x_m$ that can be approximated by a noisy version of the coarse-grid flood map. By initializing the reverse diffusion process with this new start point that is closer to the target, we effectively skip many early denoising steps. For example, in Catchment 1, we successfully reduced inference timesteps to 50 and inference time to 1 minute and 42 seconds. This strategy is inspired by latent consistency models, which accelerate diffusion-based pipelines by predicting an intermediate latent state directly, bypassing numerous iterative steps \cite{lcm}.
\end{enumerate}

\noindent Similar performance gains were observed across all catchments. The optimal number of timesteps to ensure no significant drop in performance (i.e. $<1\%$ increase in MSE) varies with the complexity of the catchment. Graphical representations of the LDM performance at different numbers of timesteps can be viewed in the Supplementary Material, along with some additional analysis on results observed in Catchment 2. Overall, the two optimizations resulted in inference speed improvements in the LDMs of up to fivefold.

We can compare the final speed-up ratio of the proposed methodology with the standard fine-grid hydrodynamic model on Catchment 1. Using a standard computing setup (Intel i5 1.90 GHz processor, 16 GB RAM, 12 solver cores), the coarse-grid flood map was generated in 5 minutes and 12 seconds, while the fine-grid flood map required 7 hours, 46 minutes, and 18 seconds to produce \cite{sgunet}. The preprocessing required to convert the coarse-grid flood map into an appropriate input format for the LDM took 2 minutes and 53 seconds. Using 50 inference timesteps and the noisy coarse-grid flood map as the starting point, the inference process of the LDM was completed in 9 minutes and 10 seconds. This results in a total generation time of 17 minutes and 15 seconds for the super-resolution flood maps. Compared to the fine-grid simulation time, this yields an approximate speed-up ratio of 27$\times$.

It is important to note that this speed-up ratio does not account for the time required to train the LDM, which is a one-time computational cost incurred prior to operational deployment. For Catchment 1, the LDM was trained over 400,000 steps on a NVIDIA H100 GPU (96 GB) across 3 days and 5 hours. While inference speed is critical for real-time applications, training time remains a significant operational constraint. In many practical scenarios, although flood maps and DEM data may be available, there is insufficient time or computational capacity to retrain a model from scratch for each new region. Under such conditions, zero-shot generalization to unseen catchments can sometimes provide useful predictions, but often falls short of the accuracy needed for reliable operational use. The alternative, relying solely on coarse-grid flood maps, is generally inadequate for real-world applications. Consequently, transfer learning offers a practical compromise: it substantially improves zero-shot performance while avoiding the prohibitive cost of training from scratch. By fine-tuning a pre-trained model on data from the new catchment, the model can rapidly adapt to new geographical contexts, delivering acceptable accuracy in a limited training window.

\subsection{Transfer Learning}
LDMs were initially trained on specific catchments to learn the flood map representation, and subsequently fine-tuned on new catchments using transfer learning for 50,000 steps. Their performance was then compared against the baseline LDMs, which were trained from scratch for 300,000 steps on their respective catchments without transfer learning.

\begin{table*}[t]
    \centering
    \begin{tabular}{>{\rowmac}c>{\rowmac}c>{\rowmac}c>{\rowmac}c>{\rowmac}c<{\clearrow}}
        \toprule
        Representation Learning Catchment & Finetuning Catchment & CG-FG MSE (cm$^2$) & SR-FG MSE (cm$^2$) & \% change $\downarrow$ \\
        \midrule
        3 & - & 158.7 & 16.8 & -89.40 \\
        1 & 3 & 158.7 & 25.9 & -83.71 \\
        2 & 3 & 158.7 & 27.5 & -82.70 \\
        \bottomrule
    \end{tabular}
    \caption{LDM performance after fine-tuning on Catchment 3 for 50,000 steps. The first row shows the baseline LDM trained for 300,000 steps without transfer learning. Transfer-learned LDMs achieve MSE reductions close to the baseline while using only one-sixth of the training steps, demonstrating the effectiveness of transfer learning in our architecture.}
    \label{tab:chowilla_transfer_learning}
\end{table*}

\Cref{tab:chowilla_transfer_learning} compares the performance of LDMs that underwent finetuning on Catchment 3 against the original Catchment 3 LDM. Similar results were obtained for the other two catchments and recorded in the Supplementary Material, indicating that the models trained via transfer learning achieve performance levels that closely approach those of the baseline LDMs despite being trained for only one-sixth the number of steps. This transfer learning process took approximately 9 hours, which is a manageable one-time overhead in operational settings and significantly less demanding than training from scratch. These findings highlight the adaptability of the LDM architecture and reinforce the earlier conclusion that diffusion-based models exhibit strong generalizability, making them well-suited for flood mapping applications across diverse geographical regions. For cases where zero-shot performance is insufficient, transfer learning provides a practical pathway to ensure accuracy and timeliness in real-world flood mapping applications. 

\subsection{Flood Inundation Analysis}
\begin{table*}[t]
    \centering
    \begin{tabular}{c|ccc|ccc|ccc}
        \toprule
        Catchment & \shortstack{CG-FG\\POD} & \shortstack{SR-FG\\POD} & \shortstack{\% change\\(POD) $\uparrow$} & \shortstack{CG-FG\\RFA} & \shortstack{SR-FG\\RFA} & \shortstack{\% change\\(RFA) $\downarrow$} & \shortstack{CG-FG\\CSI} & \shortstack{SR-FG\\CSI} & \shortstack{\% change\\(CSI) $\uparrow$} \\
        \midrule
        1 & 0.905 & 0.966 & +6.18 & 0.094 & 0.015 & -7.94 & 0.827 & 0.953 & +12.57 \\
        2 & 0.960 & 0.959 & -0.03 & 0.201 & 0.028 & -17.30 & 0.773 & 0.934 & +16.04 \\
        3 & 0.982 & 0.988 & +0.64 & 0.097 & 0.011 & -8.61 & 0.889 & 0.978 & +8.93 \\
        \bottomrule
    \end{tabular}
    \caption{POD, RFA, and CSI of LDM at the 30 cm threshold. While POD gains were minimal for Catchments 2 and 3 due to already accurate coarse-grid predictions, the LDM substantially reduced false alarms and improved CSI across all catchments, enhancing the coarse-grid flood maps.}
    \label{tab:inundation_analysis}
\end{table*}

Another avenue of model performance evaluation is the analysis of flood inundation maps, which are binary representations of flood extent derived from continuous flood depth maps. These maps play a vital operational role in guiding resource allocation during flood events \cite{fraehr_24}. Pixels exceeding a specified flood depth threshold are classified as flooded, while those below the threshold are considered dry. 

\begin{equation}
    \label{eqn:pod}
    \text{Probability of detection (POD)} = \frac{TP}{TP + FN}
\end{equation}
\begin{equation}
    \label{eqn:rfa}
    \text{Rate of false alarms (RFA)} = \frac{FP}{TP + FP}
\end{equation}
\begin{equation}
    \label{eqn:csi}
    \text{Critical success index (CSI)} = \frac{TP}{TP + FN + FP}
\end{equation}

To assess model performance, three standard metrics are computed from these inundation maps: the Probability of Detection (POD) (\cref{eqn:pod}), Rate of False Alarms (RFA) (\cref{eqn:rfa}), and Critical Success Index (CSI) (\cref{eqn:csi}) \cite{csi}. In these equations, TP, FP and FN refer to true positives, false positives and false negatives respectively. These metrics provide a comprehensive evaluation of the model’s capability to accurately delineate inundated areas, balancing detection sensitivity against false alarm rates. The POD measures the proportion of correctly identified flooded pixels, while the RFA quantifies the frequency of incorrect flood predictions. The CSI offers an overall accuracy metric, integrating both POD and RFA to evaluate practical reliability.

We compared the performance of the LDM and the coarse-grid hydrodynamic simulation in \cref{tab:inundation_analysis}. While coarse-grid simulations tend to overpredict flooding, resulting in high POD but also elevated RFA, the LDM substantially reduces false positives while maintaining comparable or superior flood detection rates. This improvement in both detection accuracy and reliability is reflected in consistently higher CSI values across all catchments. These findings underscore the effectiveness of the proposed diffusion-based approach in accurately identifying inundated regions while minimizing operational false alarms.

\subsection{DEM Ablation Study}
\label{sec:dem_ablation}

\begin{table}[t]
    \centering
    \begin{tabular}{cccc}
        \toprule
        DEM & \shortstack{CG-FG MSE} & \shortstack{SR-FG MSE} & \shortstack{\% change $\downarrow$} \\
        \midrule
        $\checkmark$ & \textbf{344.2} & \textbf{33.7} & \textbf{-90.20} \\
        $\mathbf{\times}$ & 344.2 & 70.4 & -79.56 \\
        \bottomrule
    \end{tabular}
    \caption{DEM ablation study on Catchment 1. Including the DEM reduces SR–FG MSE by over 50\%, substantially improving model accuracy.}
    \label{tab:dem_ablation}
\end{table}

To evaluate the contribution of the DEM, we conducted an ablation study using two LDMs on Catchment 1 with identical hyperparameters and architectures besides the DEM channels. \Cref{tab:dem_ablation} shows that incorporating the DEM as a conditioning signal significantly enhanced performance.

The study demonstrates that grounding the flood mapping LDM in hydrological principles fundamentally enhances its interpretability. The training regimen, which begins with physically bounded coarse-grid simulations, inherently respects laws such as volume conservation, a key metric for physical realism. The model timestep parameter is explicitly derived from the time of runoff concentration, a direct function of the size of the basin, providing a clear and interpretable mechanism for integrating the characteristics of the basin. Additionally, the integration of DEM is critical in determining flow direction and accumulation, underscoring the model's physical consistency. Each component of our framework is not merely a learned feature but an interpretable parameter with a distinct physical justification that aligns with established hydrological understanding.

\section{Conclusion}
\label{sec:conclusion}
In this paper, we proposed a novel approach that leverages diffusion models to perform super-resolution on coarse-grid flood maps, with the objective of achieving the accuracy of fine-grid flood maps while significantly reducing inference time. Our experimental results demonstrate that latent diffusion models can substantially decrease the computational time required to produce high-fidelity flood maps without compromising on accuracy. Furthermore, we have shown that diffusion-based architectures exhibit superior generalizability compared to conventional fully convolutional networks, and we have highlighted the effectiveness of transfer learning in expediting the adaptation process to new catchments. Finally, by incorporating physics-informed inputs into the model, our approach addresses the common limitation of black-box behavior in machine learning, thereby enhancing interpretability. This characteristic renders the proposed method particularly well-suited for critical applications such as disaster response and emergency planning.

\section{Acknowledgements}
This work was supported by the University of Sydney --- National University of Singapore Ignition Grants.

\newpage
{
    \small
    \bibliographystyle{ieeenat_fullname}
    \bibliography{main}

@String(PAMI = {IEEE Trans. Pattern Anal. Mach. Intell.})

@String(CVPR= {IEEE Conf. Comput. Vis. Pattern Recog.})

@String(NIPS= {Adv. Neural Inform. Process. Syst.})

@String(CVPRW= {IEEE Conf. Comput. Vis. Pattern Recog. Worksh.})

@String(NNLS = {IEEE Trans. Neural Netw. Learn. Syst.})

@String(PAMI  = {IEEE TPAMI})

@String(CVPR  = {CVPR})

@String(NIPS  = {NeurIPS})

@String(CVPRW= {CVPRW})

@techreport{gar2022,
    author = {{United Nations Office for Disaster Risk Reduction}},
    title = "Global Assessment Report on Disaster Risk Reduction 2022: Our World at Risk: Transforming Governance for a Resilient Future",
    institution = {{United Nations Office for Disaster Risk Reduction}},
    address = "Geneva",
    type = "Report",
    year = "2022"
}

@article{jodhani_23,
    title = {A review on analysis of flood modelling using different numerical models},
    journal = {Materials Today: Proceedings},
    volume = {80},
    pages = {3867-3876},
    year = 2023,
    month = apr,
    issn = {2214-7853},
    doi = {https://doi.org/10.1016/j.matpr.2021.07.405},
    url = {https://www.sciencedirect.com/science/article/pii/S221478532105269X},
    author = {Keval H Jodhani and Dhruvesh Patel and N. Madhavan},
}

@article{jafarzadegan_23,
    author = {Jafarzadegan, Keighobad and Moradkhani, Hamid and Pappenberger, Florian and Moftakhari, Hamed and Bates, Paul and Abbaszadeh, Peyman and Marsooli, Reza and Ferreira, Celso and Cloke, Hannah L. and Ogden, Fred and Duan, Qingyun},
    title = {Recent Advances and New Frontiers in Riverine and Coastal Flood Modeling},
    journal = {Reviews of Geophysics},
    volume = {61},
    number = {2},
    doi = {https://doi.org/10.1029/2022RG000788},
    url = {https://agupubs.onlinelibrary.wiley.com/doi/abs/10.1029/2022RG000788},
    eprint = {https://agupubs.onlinelibrary.wiley.com/doi/pdf/10.1029/2022RG000788},
    year = {2023},
    month = may
}

@article{zahura_22,
    title = {Predicting combined tidal and pluvial flood inundation using a machine learning surrogate model},
    journal = {Journal of Hydrology: Regional Studies},
    volume = "41",
    year = {2022},
    month = jun,
    issn = {2214-5818},
    doi = {https://doi.org/10.1016/j.ejrh.2022.101087},
    url = {https://www.sciencedirect.com/science/article/pii/S2214581822001008},
    author = {Faria T. Zahura and Jonathan L. Goodall},
}

@article{mosavi_18,
    title = {Flood Prediction Using Machine Learning Models: Literature Review},
    volume = {10},
    issn = {2073-4441},
    url = {http://dx.doi.org/10.3390/w10111536},
    doi = {10.3390/w10111536},
    number = {11},
    journal = {Water},
    publisher = {MDPI AG},
    author = {Mosavi, Amir and Ozturk, Pinar and Chau, Kwok-wing},
    year = {2018},
    month = oct
}

@article{fraehr_24,
    title = {Assessment of surrogate models for flood inundation: The physics-guided LSG model vs. state-of-the-art machine learning models},
    journal = {Water Research},
    volume = {252},
    year = {2024},
    month = mar,
    day = 15,
    issn = {0043-1354},
    doi = {https://doi.org/10.1016/j.watres.2024.121202},
    url = {https://www.sciencedirect.com/science/article/pii/S0043135424001027},
    author = {Niels Fraehr and Quan J. Wang and Wenyan Wu and Rory Nathan},
}

@conference{ddpm,
    author = {Ho, Jonathan and Jain, Ajay and Abbeel, Pieter},
    title = {Denoising diffusion probabilistic models},
    year = {2020},
    month=dec,
    isbn = {9781713829546},
    publisher = {Curran Associates Inc.},
    address = {Red Hook, NY, USA},
    booktitle = NIPS,
    articleno = {574},
    numpages = {12},
    pages = {6840-6851},
    location = {Vancouver, BC, Canada},
    series = {NIPS '20}
}

@misc{improved_ddpm,
    title = {Improved Denoising Diffusion Probabilistic Models}, 
    author = {Alex Nichol and Prafulla Dhariwal},
    year = {2021},
    month = feb,
    eprint = {2102.09672},
    archivePrefix = {arXiv},
    primaryClass = {cs.LG},
    url = {https://arxiv.org/abs/2102.09672}, 
}

@article{kazerouni_23,
    title = {Diffusion models in medical imaging: A comprehensive survey},
    journal = {Medical Image Analysis},
    volume = {88},
    year = {2023},
    month = aug,
    issn = {1361-8415},
    doi = {https://doi.org/10.1016/j.media.2023.102846},
    url = {https://www.sciencedirect.com/science/article/pii/S1361841523001068},
    author = {Amirhossein Kazerouni and Ehsan Khodapanah Aghdam and Moein Heidari and Reza Azad and Mohsen Fayyaz and Ilker Hacihaliloglu and Dorit Merhof}
}

@article{croitoru_23,
    title = {Diffusion Models in Vision: A Survey},
    volume = {45},
    issn = {1939-3539},
    url = {http://dx.doi.org/10.1109/TPAMI.2023.3261988},
    doi = {10.1109/tpami.2023.3261988},
    number = {9},
    journal = PAMI,
    publisher = {Institute of Electrical and Electronics Engineers (IEEE)},
    author = {Croitoru, Florinel-Alin and Hondru, Vlad and Ionescu, Radu Tudor and Shah, Mubarak},
    year = {2023},
    month = sep, 
    pages = {10850–10869}
}

@article{mudashiru_21,
    title = {Flood hazard mapping methods: A review},
    journal = {Journal of Hydrology},
    volume = {603},
    year = {2021},
    month = dec,
    issn = {0022-1694},
    doi = {https://doi.org/10.1016/j.jhydrol.2021.126846},
    url = {https://www.sciencedirect.com/science/article/pii/S0022169421008969},
    author = {Rofiat Bunmi Mudashiru and Nuridah Sabtu and Ismail Abustan and Waheed Balogun}
}

@article{bhattarai_24,
    title = {Rapid prediction of urban flooding at street-scale using physics-informed machine learning-based surrogate modeling},
    journal = {Total Environment Advances},
    volume = {12},
    year = {2024},
    month = dec,
    issn = {2950-3957},
    doi = {https://doi.org/10.1016/j.teadva.2024.200116},
    url = {https://www.sciencedirect.com/science/article/pii/S2950395724000213},
    author = {Yogesh Bhattarai and Sunil Bista and Rocky Talchabhadel and Sunil Duwal and Sanjib Sharma},
}

@manual{hecras,
    title = {{HEC-RAS} Hydraulic Reference Manual},
    author = "Gary W. Brunner",
    organization = "Hydrologic Engineering Center, Institute for Water Resources, U.S. Army Corps of Engineers",
    month = dec,
    year = {2020},
    url = {https://www.hec.usace.army.mil/confluence/rasdocs/ras1dtechref/latest},
}

@article{kumar_23,
    author = {Kumar, Vijendra and Sharma, Kul Vaibhav and Caloiero, Tommaso and Mehta, Darshan J. and Singh, Karan},
    title = {Comprehensive Overview of Flood Modeling Approaches: A Review of Recent Advances},
    journal = {Hydrology},
    volume = {10},
    year = {2023},
    month = jun,
    number = {7},
    url = {https://www.mdpi.com/2306-5338/10/7/141},
    issn = {2306-5338},
    doi = {10.3390/hydrology10070141}
}

@article{karim_23,
    author = {Karim, Fazlul and Armin, Mohammed Ali and Ahmedt-Aristizabal, David and Tychsen-Smith, Lachlan and Petersson, Lars},
    title = {A Review of Hydrodynamic and Machine Learning Approaches for Flood Inundation Modeling},
    journal = {Water},
    volume = {15},
    year = {2023},
    month = feb,
    number = {3},
    url = {https://www.mdpi.com/2073-4441/15/3/566},
    issn = {2073-4441},
    doi = {10.3390/w15030566}
}

@article{bentivoglio_22,
    author = {Roberto Bentivoglio and Elvin Isufi and Sebastian Nicolaas Jonkman and Riccardo Taormina},
    title = {Deep learning methods for flood mapping: a review of existing applications and future research directions},
    journal = {Hydrology and Earth System Sciences},
    volume = {26},
    year = {2022},
    month = aug,
    day = {25},
    number = {16},
    pages = {4345-4378},
    url = {https://hess.copernicus.org/articles/26/4345/2022/},
    doi = {10.5194/hess-26-4345-2022}
}

@article{tavus_22,
    author = {Beste Tavus and Recep Can and Sultan Kocaman},
    title = {{A CNN-based flood mapping approach using Sentinel-1 data}},
    journal = {ISPRS Annals of the Photogrammetry, Remote Sensing and Spatial Information Sciences},
    volume = {V-3-2022},
    year = {2022},
    month = may,
    day = 17,
    pages = {549--556},
    url = {https://isprs-annals.copernicus.org/articles/V-3-2022/549/2022/},
    doi = {10.5194/isprs-annals-V-3-2022-549-2022}
}

@article{xiang_20,
    author = {Xiang, Zhongrun and Yan, Jun and Demir, Ibrahim},
    title = {A Rainfall-Runoff Model With {LSTM}-Based Sequence-to-Sequence Learning},
    journal = {Water Resources Research},
    volume = {56},
    number = {1},
    doi = {https://doi.org/10.1029/2019WR025326},
    url = {https://agupubs.onlinelibrary.wiley.com/doi/abs/10.1029/2019WR025326},
    eprint = {https://agupubs.onlinelibrary.wiley.com/doi/pdf/10.1029/2019WR025326},
    year = {2020},
    month = jan,
    day = 3,
}

@article{murdoch_19,
    author = {W. James Murdoch  and Chandan Singh  and Karl Kumbier  and Reza Abbasi-Asl  and Bin Yu },
    title = {Definitions, methods, and applications in interpretable machine learning},
    journal = {Proceedings of the National Academy of Sciences},
    volume = {116},
    number = {44},
    pages = {22071-22080},
    year = {2019},
    month = oct,
    day = 16,
    doi = {10.1073/pnas.1900654116},
    URL = {https://www.pnas.org/doi/abs/10.1073/pnas.1900654116},
    eprint = {https://www.pnas.org/doi/pdf/10.1073/pnas.1900654116},
}

@article{yang_24,
    title = {Rapid urban flood inundation forecasting using a physics-informed deep learning approach},
    journal = {Journal of Hydrology},
    volume = {643},
    year = {2024},
    month = nov,
    issn = {0022-1694},
    doi = {https://doi.org/10.1016/j.jhydrol.2024.131998},
    url = {https://www.sciencedirect.com/science/article/pii/S0022169424013945},
    author = {Fang Yang and Wu Ding and Jianshi Zhao and Lixiang Song and Dawen Yang and Xudong Li},
}

@article{yin_24,
    title = {Fast high-fidelity flood inundation map generation by super-resolution techniques},
    author = {Yin, Zeda and Saadati, Yasaman and Hu, Beichao and Leon, Arturo S and Amini, M Hadi and McDaniel, Dwayne},
    journal = {Journal of Hydroinformatics},
    volume = {26},
    number = {1},
    pages = {319--336},
    year = {2024},
    month = jan,
    publisher = {IWA Publishing}
}

@conference{ronneberger_15,
    author = "Ronneberger, Olaf and Fischer, Philipp and Brox, Thomas",
    editor = "Navab, Nassir and Hornegger, Joachim and Wells, William M. and Frangi, Alejandro F.",
    title = "U-Net: Convolutional Networks for Biomedical Image Segmentation",
    booktitle = "Medical Image Computing and Computer-Assisted Intervention -- MICCAI 2015",
    year = "2015",
    month = oct,
    publisher = "Springer International Publishing",
    address = "Munich, Germany",
    pages = "234--241",
    isbn = "978-3-319-24574-4"
}

@conference{hu_19,
    author = {Hu, Xiaodan and Naiel, Mohamed A. and Wong, Alexander and Lamm, Mark and Fieguth, Paul},
    booktitle = CVPRW, 
    title = {{RUNet}: A Robust {UNet} Architecture for Image Super-Resolution}, 
    year = {2019},
    month = jun,
    pages = {505-507},
    address = {Long Beach, CA, USA},
    doi = {10.1109/CVPRW.2019.00073}
}

@article{moser_24,
    title = {Diffusion Models, Image Super-Resolution, and Everything: A Survey},
    issn = {2162-2388},
    url = {http://dx.doi.org/10.1109/TNNLS.2024.3476671},
    doi = {10.1109/tnnls.2024.3476671},
    journal = NNLS,
    publisher = {Institute of Electrical and Electronics Engineers (IEEE)},
    author = {Moser, Brian B. and Shanbhag, Arundhati S. and Raue, Federico and Frolov, Stanislav and Palacio, Sebastian and Dengel, Andreas},
    year = {2024},
    month = jun,
    pages = {1–21} 
}

@conference{palette,
    author = {Saharia, Chitwan and Chan, William and Chang, Huiwen and Lee, Chris and Ho, Jonathan and Salimans, Tim and Fleet, David and Norouzi, Mohammad},
    title = {Palette: Image-to-Image Diffusion Models},
    year = {2022},
    month = aug,
    isbn = {9781450393379},
    publisher = {Association for Computing Machinery},
    address = {New York, NY, USA},
    url = {https://doi.org/10.1145/3528233.3530757},
    doi = {10.1145/3528233.3530757},
    booktitle = {ACM SIGGRAPH 2022 Conference Proceedings},
    articleno = {15},
    numpages = {10},
    location = {Vancouver, BC, Canada},
    series = {SIGGRAPH '22}
}

@article{sr3,
    author = {Saharia, Chitwan and Ho, Jonathan and Chan, William and Salimans, Tim and Fleet, David J. and Norouzi, Mohammad},
    journal = PAMI, 
    title = {Image Super-Resolution via Iterative Refinement}, 
    year = {2023},
    month = sep,
    day = 12,
    volume = {45},
    number = {4},
    pages = {4713-4726},
    doi = {10.1109/TPAMI.2022.3204461}
}

@conference{rombach_22,
    author = {Rombach, Robin and Blattmann, Andreas and Lorenz, Dominik and Esser, Patrick and Ommer, Bjorn},
    booktitle = CVPR,
    title = {High-Resolution Image Synthesis with Latent Diffusion Models},
    year = {2022},
    pages = {10674-10685},
    doi = {10.1109/CVPR52688.2022.01042},
    url = {https://doi.ieeecomputersociety.org/10.1109/CVPR52688.2022.01042},
    publisher = {IEEE Computer Society},
    address = {Los Alamitos, CA, USA},
    month = jun
}

@article{sgunet,
    title = {Subgrid informed neural networks for high-resolution flood mapping},
    journal = {Journal of Hydrology},
    volume = {660},
    year = {2025},
    month = oct,
    issn = {0022-1694},
    doi = {https://doi.org/10.1016/j.jhydrol.2025.133329},
    url = {https://www.sciencedirect.com/science/article/pii/S0022169425006675},
    author = {Herath Mudiyanselage Viraj Vidura Herath and Lucy Marshall and Abhishek Saha and Sanka Rasnayaka and Sachith Seneviratne},
}

@article{bomers_19,
    title = "The influence of grid shape and grid size on hydraulic river modelling performance",
    author = "Anouk Bomers and Schielen, Ralph Mathias Johannes and Hulscher, Suzanne J. M. H.",
    year = "2019",
    month = oct,
    doi = "10.1007/s10652-019-09670-4",
    language = "English",
    volume = "19",
    pages = "1273--1294",
    journal = "Environmental Fluid Mechanics",
    issn = "1567-7419",
    publisher = "Springer Science and Business Media B.V.",
    number = "5",
}

@article{csi,
    title = {The critical success index as an indicator of Warning skill},
    author = {Joseph T. Schaefer},
    journal = {Weather and Forecasting},
    year = {1990},
    month = dec,
    volume = {5},
    pages = {570-575},
    url = {https://api.semanticscholar.org/CorpusID:122542836}
}

@article{he_23,
    title = {Deep learning enables super-resolution hydrodynamic flooding process modeling under spatiotemporally varying rainstorms},
    journal = {Water Research},
    volume = {239},
    year = {2023},
    month = jul,
    issn = {0043-1354},
    doi = {https://doi.org/10.1016/j.watres.2023.120057},
    url = {https://www.sciencedirect.com/science/article/pii/S0043135423004931},
    author = {Jian He and Limin Zhang and Te Xiao and Haojie Wang and Hongyu Luo},
}

@article{song_25,
    author = {Wenke Song and Mingfu Guan and Kaihua Guo and Dapeng Yu},
    title = {Rapid flood inundation mapping by integrating deep learning-based image super-resolution with coarse-grid hydrodynamic modeling},
    journal = {Engineering Applications of Computational Fluid Mechanics},
    volume = {19},
    number = {1},
    year = {2025},
    month = mar,
    publisher = {Taylor \& Francis},
    doi = {10.1080/19942060.2025.2481115},
    url = {https://doi.org/10.1080/19942060.2025.2481115},
}

@article{choi_25,
    title = {{FLO-SR: Deep learning-based urban flood super-resolution model}},
    journal = {Journal of Hydrology},
    volume = {661},
    year = {2025},
    month = nov,
    issn = {0022-1694},
    doi = {https://doi.org/10.1016/j.jhydrol.2025.133529},
    url = {https://www.sciencedirect.com/science/article/pii/S0022169425008674},
    author = {Hyeonjin Choi and Hyuna Woo and Minyoung Kim and Hyungon Ryu and Jun-Hak Lee and Seungsoo Lee and Seong Jin Noh},
}

@article{ding_20,
    title = {{Interpretable spatio-temporal attention LSTM model for flood forecasting}},
    journal = {Neurocomputing},
    volume = {403},
    pages = {348-359},
    year = {2020},
    month = aug,
    issn = {0925-2312},
    doi = {https://doi.org/10.1016/j.neucom.2020.04.110},
    url = {https://www.sciencedirect.com/science/article/pii/S0925231220307530},
    author = {Yukai Ding and Yuelong Zhu and Jun Feng and Pengcheng Zhang and Zirun Cheng},
}

@article{cache_24,
    author = {Tabea Cache and Milton Salvador Gomez and Tom Beucler and Jovan Blagojevic and Jo{\~a}o Paulo Leitao and Nadav Peleg},
    title = {Enhancing generalizability of data-driven urban flood models by incorporating contextual information},
    journal = {Hydrology and Earth System Sciences},
    volume = {28},
    year = {2024},
    month = dec,
    number = {24},
    pages = {5443--5458},
    url = {https://hess.copernicus.org/articles/28/5443/2024/},
    doi = {10.5194/hess-28-5443-2024}
}

@misc{lcm,
    title = {Latent Consistency Models: Synthesizing High-Resolution Images with Few-Step Inference}, 
    author = {Simian Luo and Yiqin Tan and Longbo Huang and Jian Li and Hang Zhao},
    year = {2023},
    month = oct,
    eprint = {2310.04378},
    archivePrefix = {arXiv},
    primaryClass = {cs.CV},
    url = {https://arxiv.org/abs/2310.04378}, 
}
}

\end{document}